\documentclass[final]{cvpr}

\usepackage{times}
\usepackage{epsfig}
\usepackage{graphicx}
\usepackage{amsmath}
\usepackage{amssymb}
\usepackage{multirow}
\usepackage{chapterbib}
\usepackage{titling}

\usepackage[pagebackref=true,breaklinks=true,colorlinks,bookmarks=false]{hyperref}

\def\cvprPaperID{2983} 
\def\confYear{CVPR 2021}

\begin{document}

\font\elvbfx  = ptmb scaled 1500
\title{\elvbfx{CRFace: Confidence Ranker for Model-Agnostic Face Detection Refinement}}

\author{Noranart Vesdapunt\\
Microsoft Cloud\&AI	\\
{\tt\small noves@microsoft.com}
\and
Baoyuan Wang\\
Xiaobing.AI\\
{\tt\small zjuwby@gmail.com	}
}
\date{\vspace{-2ex}}

\maketitle

\begin{abstract}
Face detection is a fundamental problem for many downstream face applications, and there is a rising demand for faster, more accurate yet support for higher resolution face detectors. Recent smartphones can record a video in 8K resolution, but many of the existing face detectors still fail due to the anchor size and training data. We analyze the failure cases and observe a large number of correct predicted boxes with incorrect confidences. To calibrate these confidences, we propose a confidence ranking network with a pairwise ranking loss to re-rank the predicted confidences locally within the same image. Our confidence ranker is model-agnostic, so we can augment the data by choosing the pairs from multiple face detectors during the training, and generalize to a wide range of face detectors during the testing. On WiderFace, we achieve the highest AP on the single-scale, and our AP is competitive with the previous multi-scale methods while being significantly faster. On 8K resolution, our method solves the GPU memory issue and allows us to indirectly train on 8K. We collect 8K resolution test set to show the improvement, and we will release our test set as a new benchmark for future research.

\end{abstract}

\begin{figure*}[t]
\centering
\includegraphics[width=1.0\textwidth]{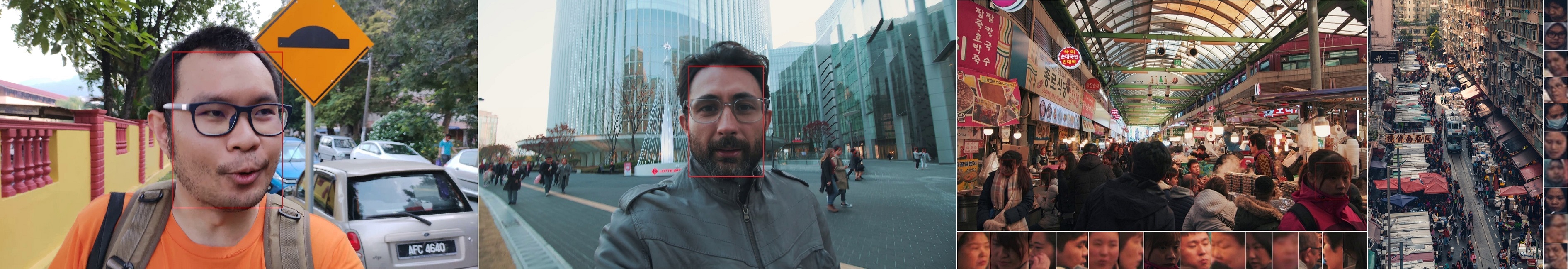}
\caption{(Best viewed electronically) Examples of our 8K test set. The first two images contain both large and small faces. Detecting on the large resolution will fail the large face due to the small anchor size and the training data. Down-sampling will cause the small faces to be too small to detect. Last two images contain a large amount of small faces that are expensive to label.}
\label{fig:8k_example}
\end{figure*}

\vspace{-15pt}
\section{Introduction}
Face detection is a long-standing research topic in computer vision. Many important downstream applications need to build on top of it, including face reconstruction, face tracking and face recognition, etc. Thanks to Convolution Neural Network, face detection has been improved significantly in the past few years. However, as the camera hardware is evolving, the demand for an even faster, more accurate, and support for higher resolution in face detector is rising. Recent smartphones (Samsung Galaxy S20 Ultra, Xiaomi MI 10 Pro) can capture a 108MP image and record a video in 8K resolution. We believe that 8K resolution will be practical in the near future, but a lot of existing works \cite{HAMBox,Deng_2020_CVPR,li2019dsfd,zhang2017s3fd} still fail on many 8K resolution inputs.

The main reasons for failure are anchor size and training data. Most of the anchors are designed for the most popular face detection dataset, WiderFace \cite{yang2016wider} which only has an image width of 1024. The common largest anchor size is 512 \cite{HAMBox,Deng_2020_CVPR,li2019dsfd,zhang2017s3fd}, so the detector has to predict up to 7K residuals. While predicting such large residuals is hard, but possible, the classifier has never seen such large resolution in the training set and it will almost always suppress all these boxes. Collecting 8K training data is expensive because a single image could contain hundreds of faces (Fig. \ref{fig:8k_example}). Designing a large anchor requires the network to be deep enough to output multiple scales up to 8K, but BFBox \cite{Liu_2020_CVPR} shows that the average precision (AP) decreases when the network becomes deeper. While down-sampling the image allows the detector to find a large face, the small face could be reduced to just 1 pixel (Fig. \ref{fig:8k_example}). We also found that 48GB GPU memory is not enough to train RetinaFace \cite{Deng_2020_CVPR} on 8K resolution. As there are many challenges in 8K face detection, multi-scale \cite{hu2017finding} is perhaps the most reasonable workaround by predicting the small faces in 8K, and the large faces in the smaller resolution, then fuse them together with box voting \cite{Gidaris_2015_ICCV}. However, this approach still relies on the correct confidence as the majority of the high confidence boxes will determine the result. Multi-scale also decreases the 8K prediction speed (which is already severely slow) to become much slower than a single-scale.

\begin{figure*}[t]
\centering
\includegraphics[width=1.0\textwidth]{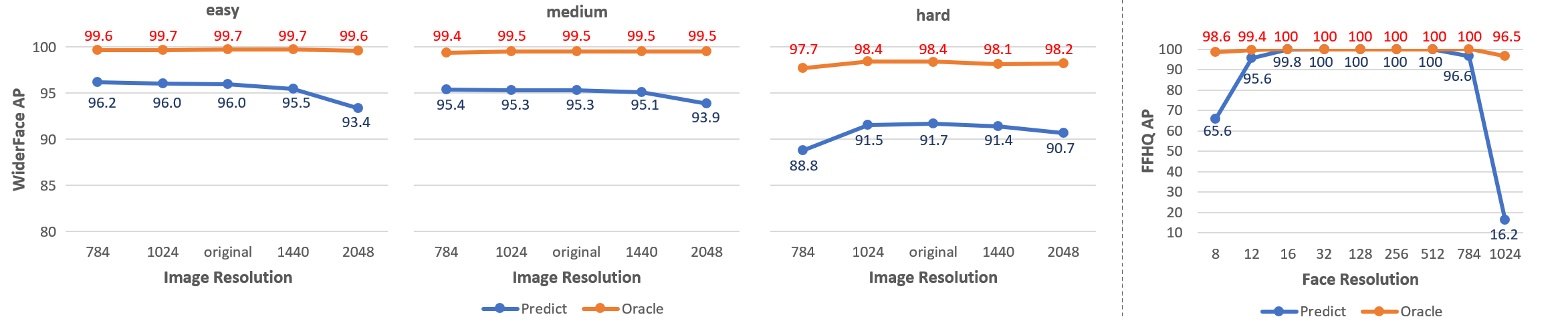}
\caption{AP of predicted vs oracle box from HRNet \cite{WangSCJDZLMTWLX19,SunXLW19} face detector. We resize the image height of WiderFace validation set (left figure) to multiple resolutions (x-axis), test on the single-scale, and observe a large consistent gap across all resolutions. We double verify the gap in a more controlled test set by labeling 1,875 images from FFHQ dataset (right figure). We tested on resized face resolution and the conclusion remains the same. This shows that most of the corrected boxes are already predicted, but the confidences are incorrect.}
\label{fig:oracle}
\end{figure*}

We analyze the fail cases and found that most of the correct box locations have already existed because many detectors predict a large number of boxes, but the confidences are very low. To systematically test these boxes, we replace the prediction confidence with the intersect-over-union (IoU) between the predicted box and the closest ground truth box and call it: oracle confidence. Fig. \ref{fig:oracle} shows that the oracle prediction has a consistently high AP on WiderFace validation set across multiple resolutions, and by testing only on a single-scale, it can outperform the state-of-the-art \cite{HAMBox} by a large margin (AP Hard: 93.3\% vs 98.4\%). On our collected FFHQ \cite{Karras_2019_CVPR} dataset, AP could even be increased from 16.2\% to 96.5\%. We saw the possibility of closing the gap between the predicted and oracle confidence. Since we can use the oracle confidence as ground truth, we formulate this confidence refinement problem as supervised learning.

We initially try to learn a regressor that takes an image and face detector's output as input to predict the new confidence, then pass them into NMS. This, however, does not improve AP because regressing to the exact floating point is challenging. The confidences of the boxes around the face boundary could be only $10^{-4}$ different from each other, and a very small regression error can significantly change the confidence order, and thus changing the outcome of the non-maximal suppression (NMS). Since NMS is a greedy technique that is only affected by the order of the confidence, we propose to relax the problem from regression to local ordinal ranking where we only need to rank the refined confidences within the same image and we can ignore the magnitude of the confidence. We use a pairwise ranking loss to enforce such constraint, and we show that our ranker learns to preserve the order of confidence and improve AP.

\begin{figure*}[t]
\centering
\includegraphics[width=1.0\textwidth]{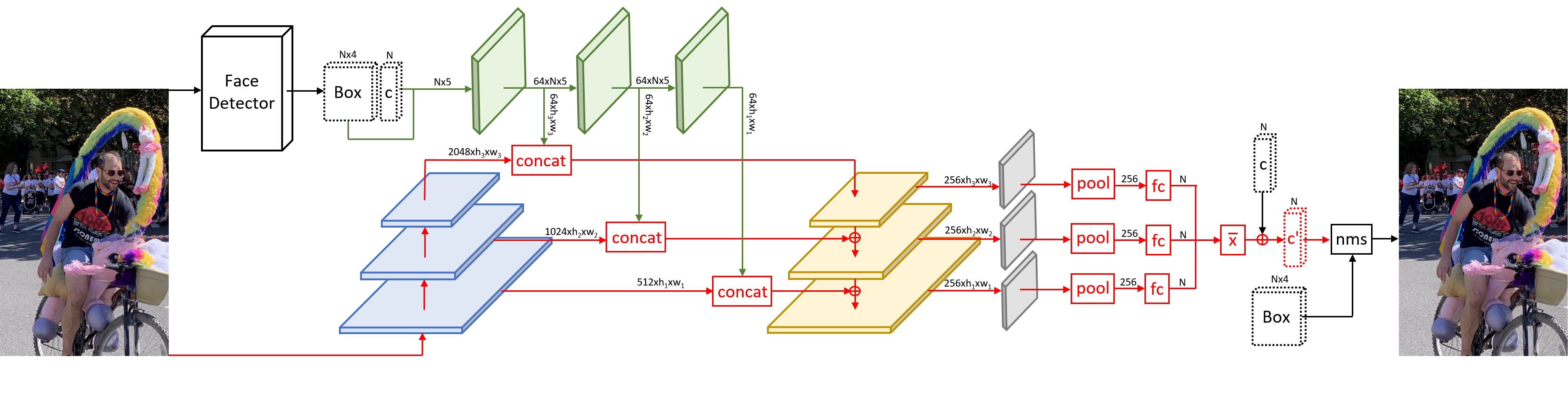}
\caption{An overview of our pipeline. We use FPN with 3 scales to extract features from the image. We pass the face detector's output into our Box Processing Network (BPN, denoted in green) and interpolate BPN's feature dimension to concatenate with the image features. Our confidence module applies global average pooling and fully-connected layer to the features from each scale, then averages them to output the refined confidence ($c'$). The final prediction is the bounding box from the face detector and the new confidence from our network.}
\label{fig:architecture}
\end{figure*}

We propose a confidence ranking network by taking the bounding box and confidence prediction from a face detector to output new refined confidences. On top of Feature Pyramid Network (FPN \cite{Lin_2017_CVPR}), we add a Box Processing Network (BPN) to extract features from the face detector's output and interpolate them to concatenate with the image feature, then pass them into our confidence module. Fig. \ref{fig:architecture} demonstrates our pipeline. We design our network to be model-agnostic, so in theory, it can be used with any object detector, as long as the outputs are bounding boxes and their corresponding confidence values, and we show the generalization on three face detectors including HAMBox \cite{HAMBox}, RetinaFace \cite{Deng_2020_CVPR}, and HRNet \cite{WangSCJDZLMTWLX19}. Model-agnostic design is important for a fair comparison with multi-scale which is the de facto post-processing for modern face detectors in order to get high AP, with the cost of slow speed. Since our network only needs a single-scale to run, we are a few times faster than those state-of-the-arts \cite{HAMBox, Deng_2020_CVPR} with multi-scale, while still retain a competitive AP on WiderFace. On a single-scale, our method is the new state-of-the-art. Furthermore, our method allows us to solve the GPU memory issue and indirectly train on 8K (it would otherwise not be possible even with 48GB GPU memory) by backpropagating up until the 8K prediction from the face detector. We then collect 8K test set to demonstrate the effectiveness of our method and set up a new benchmark for future camera hardware. To summarize, our main contributions are:

\begin{enumerate}
\item We propose to refine confidence in a local relative setting with our ranking loss, in contrast to existing works that regress to the absolute value in a global manner, inspired by the failure case analysis on WiderFace validation set and our collected FFHQ dataset.



\item We propose a confidence ranking network to achieve the new state-of-the-art on the single-scale face detector. Our AP is competitive with the previous multi-scale state-of-the-art \cite{HAMBox} while remaining a few times faster. Our network is model-agnostic and we show the generalization on HAMBox \cite{HAMBox}, RetinaFace \cite{Deng_2020_CVPR}, and HRNet \cite{WangSCJDZLMTWLX19} respectively.

\item Our method solves the GPU memory issue and allows us to indirectly train on 8K resolution to further increase AP. We collect 8K resolution test set to show the improvement, and we will release our test sets as a new benchmark for future camera hardware.
\end{enumerate}

\vspace{-5pt}
\section{Related Work}
\noindent{\textbf{Object Detection}} has been advanced significantly by deep learning and there are many types of deep learning based detectors. Two-stage detectors \cite{6909475,girshick2015fast,ren2015faster,8237584,cai2018cascade,NIPS2016_577ef115} generate region proposals in the first stage, then refine them in the second stage. One-stage detectors \cite{Redmon_2016_CVPR,Redmon_2017_CVPR,liu2016ssd,lin2017focal,zhang2018single} remove the first stage and run directly on the predefined anchors. Anchor-free detectors \cite{kong2019foveabox,Law_2018_ECCV,FCOS2019Tian,268e88c096014d088550b3f9b077dc48,DBLP:journals/corr/abs-1904-07850,Zhou_2019_CVPR} remove anchor to reduce design parameters. Although each type is different, we design our network to only use the box and confidence from the input detector, so in theory, our work is compatible with these wide ranges of detectors. \\

\vspace{-5pt}
\noindent{\textbf{Face Detection}} is a special case of object detection and they can leverage each other. On top of a generic object detector, past works propose to change the anchor matching \cite{zhang2017s3fd,Chi2019,ming2019group,zhu2018seeing,li2019dsfd,HAMBox} or change the anchor sampling \cite{Najibi_2019_CVPR} to detect small faces, add context module \cite{hu2017finding,tang2018pyramidbox,li2019pyramidbox++} to learn contextual features, add face alignment task \cite{7553523,Chaudhuri_2019_CVPR,Deng_2020_CVPR} to leverage different dataset, search for face-specific backbone \cite{Liu_2020_CVPR} to increase network capacity, or train progressively \cite{zhuprogressface} to handle multi-scale. While training strategies are varied, recent face detectors focus on the one-stage design \cite{zhang2017s3fd,najibi2017ssh,Chaudhuri_2019_CVPR,Deng_2020_CVPR,zhuprogressface,tang2018pyramidbox,li2019pyramidbox++,HAMBox} due to the benefit of dense anchor sampling and scale variation from feature pyramid. Our network is inspired by these works and we add extra layers to process the box and output the new confidence. Many state-of-the-arts \cite{Deng_2020_CVPR,zhuprogressface,HAMBox} consistently see improvement when applying multi-scale \cite{hu2017finding}. We found that the detector can predict most of the corrected box, but the confidences are incorrect, so we propose to speed up by refining the confidence and only run on a single-scale.     \\

\vspace{-5pt}
\noindent{\textbf{Detection Refinement}} is commonly used in object detection \cite{zhang2018single,Cai_2018_CVPR,Najibi_2016_CVPR,Gong2019ImprovingMO,Aroulanandam2019,rukhovich2020iterdet}. The idea is to detect the object multiple times to refine the prediction. 
In face detection, previous works \cite{Feng_2017_CVPR_Workshops,inbook,7553523,Luo2018} propose to iteratively regress the box, or use cascade networks \cite{8794507,7780745,rongqi,Zeng2019FastCF} to refine the prediction. Our approach is a form of refinement where we propose a second network to only refine the confidence with extra supervision from the oracle confidence. While IoU-Net \cite{Jiang_2018_ECCV} has proposed the oracle confidence for object detection before, we are the first to introduce the oracle confidence to face detection through the observation from Fig. \ref{fig:oracle}. For face detection, confidence is the bottleneck for the prediction error, so we can bypass the box refinement step in IoU-Net, which is a heavy optimization process. We also show the importance of changing from Smooth L1 loss in IoU-Net to our ranking loss. Lastly, our design is model-agnostic, in contrast to IoU-Net's dependency on the PrRoiPooling \cite{Jiang_2018_ECCV} layer.  \\

\vspace{-5pt}
\noindent{\textbf{Ranking Loss}} is a well-studied loss function, especially in information retrieval, and is grouped into Pointwise (\cite{conf/nips/LiBW07,Crammer01prankingwith,herbrich2000large}, Pairwise (\cite{burges2005learning,NIPS2006_af44c4c5,burges2010ranknet}), and Listwise (\cite{DBLP:conf/icml/CaoQLTL07,DBLP:conf/icml/XiaLWZL08,volkovs_2009}). In this paper, we show the issue with the existing regression losses, and with such a simple pairwise ranking loss, we were able to achieve significant improvement.

\section{Methodology}
We built our baseline by re-implementing HAMBox \cite{HAMBox}, and to simplify the baseline, we do not use Pyramid Anchor \cite{tang2018pyramidbox}, Deep Head \cite{lin2017focal}, or regression-aware focal loss \cite{HAMBox}. We instead change the backbone of HAMBox from ResNet-50 \cite{DBLP:conf/cvpr/HeZRS16} to HRNet-W48 \cite{WangSCJDZLMTWLX19,SunXLW19} to further close the AP gap. With the new baseline, we analyze the failure case, propose a ranking loss, and design a model-agnostic confidence ranking network to further improve AP.

\subsection{Oracle Confidence}
We first look at the failure cases from our baseline, and we found multiple reasons such as occlusion, small face, large pose, etc. While we could tackle one failure case at a time, we found a common behavior that the correct bounding boxes are predicted, but the predicted confidences are incorrect. Out of the 4,062 false negative faces on WiderFace validation set, only 531 faces can not be predicted by the detector. If we can correct these confidences, 87\% of the false negatives will be fixed, and all of the false positives will be suppressed by setting their confidences to zero.

To achieve better confidences, we first consider optimizing new confidences based on the predicted box as the input, and AP as the cost function. This optimization is too heavy as our baseline detector can predict $>$5k boxes and we need to rerun NMS and recompute AP on every iteration. By definition, the new confidence should be a distance metric between the predicted and ground truth box, so although not the most optimal solution, we replace the predicted confidence with the IoU between a predicted box and the closest ground truth box, and call it: oracle confidence. We tested our oracle prediction across multiple scales of WiderFace validation set and we see AP$>$99\% on the Easy and Medium set. On Hard set, if we resize the resolution down to 784 (almost half), we start to see the AP drop to 97.7\% because some of the faces are being resized to $<$5 pixels (could be 1 pixel). The missing boxes are the limitation of our method, but there is no alternative solution that can detect a 1-pixel face. By testing on WiderFace, one could argue that the challenging image condition causes the confidence prediction to be less accurate, and the oracle prediction improvement came from using the ground truth to overcome these issues. Therefore, we labeled 1,875 images from FFHQ, which only has high quality, single face with minimal background, to double verify our oracle confidence. Fig. \ref{fig:oracle} shows the AP comparisons.

\subsection{Ranking Loss}
After obtaining the oracle confidence, we can train to refine confidence in a supervised learning fashion. We first formulate the problem as regression and test multiple loss functions. While the regressor is powerful enough to move the new confidence towards the oracle confidence distribution, it is not accurate enough to preserve the order, especially for boxes around the face. This is due to the small IoU difference between them and it results in a very small loss. To relax this problem, we observe a few priors from NMS. First, the confidence magnitude does not change the NMS output, as long as the order remains the same within an image. Second, the order across different images does not change the outcome of NMS, which means these confidences do not have to be globally ranked across the entire dataset. We just need to focus on the order of the confidence within the same image, so we propose to use a pairwise ranking loss. We first define the relationship between the ground truth pair as a binary classification problem.

\begin{equation}
\label{eq1}
Y(c_{gt,1},c_{gt,2}) = 
\begin{cases}
0 &\mbox{if $c_{gt,1}>c_{gt,2}$}\\
1 &\mbox{if $c_{gt,1}<=c_{gt,2}$} \\
\end{cases}
\end{equation}

\noindent{where} $c_{gt}$ is the oracle confidence or the IoU between the predicted and ground truth box. $Y$ is the class label from a pair of oracle confidences. We then subtract a pair of predicted confidences to obtain a linear difference between them and pass it into a Sigmoid cross entropy loss.

\begin{multline} 
\label{eq2}
L(c',c_{gt}) = -(c'_1-c'_2)*log(Y(c_{gt,1},c_{gt,2}))\\
-(1-(c'_1-c'_2))*log(1-Y(c_{gt,1},c_{gt,2}))
\end{multline}

\begin{figure}[t]
\centering
\includegraphics[width=0.47\textwidth]{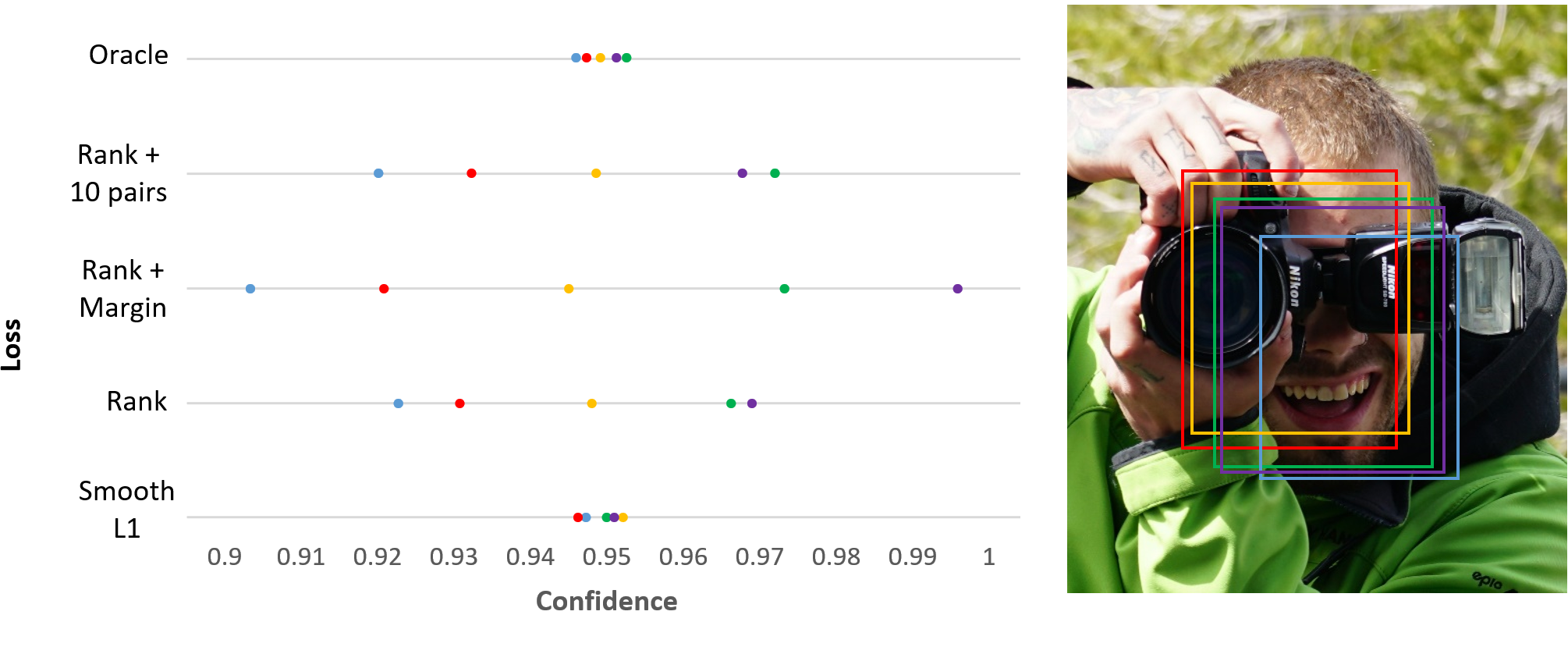}
\caption{Confidence distribution visualization for each loss function. Smooth L1 loss has the closest distribution to the oracle (ground truth), but the order is not preserved. Ranking loss does not respect the confidence magnitude, but performs better at preserving the order. Margin rescaling technique \cite{Joachims:2009:PSO:1592761.1592783,Taskar:2003:MMN:2981345.2981349} can push the distribution further but does not change the order much. Selecting the number of pairs (10 pairs) can also change the order.}
\label{fig:pred_dist}
\end{figure}

\noindent{where} $c'_1$ and $c'_2$ are the predicted confidence pair. We consider using the margin rescaling technique \cite{Joachims:2009:PSO:1592761.1592783,Taskar:2003:MMN:2981345.2981349} and this loss can push the confidences away from each other (Fig. \ref{fig:pred_dist}), but the AP is about the same as our simpler ranking loss. We select the pair by sorting the predicted confidences in descending order and pick the neighbor pair. This pair selecting strategy exposes the pair that the detector struggles with the most, but we also would like to include an easy pair as well, not just the pair with a subtle difference. We add n-pair to the original neighbor pair by skipping n confidences and divide the loss by n, for example, 2-pairs are the neighbor pair and the odd and even pair. We observe that the number of pairs can affect AP. Fig. \ref{fig:pred_dist} shows the comparison of the confidence distributions from each loss.

\subsection{Model-Agnostic Design}
We consider multiple types of inputs for our confidence ranking network. Our design principle is to make our network be compatible with as many face detectors as possible so that we can increase the number of training data by combining the predictions from all these detectors. Our model-agnostic design can be viewed as a plug-in module for an existing face detector. In this paper, we use this refinement to tackle the 8K resolution input, but the same idea can also be applied for other scenarios, and since we do not change the predicted box, we are preserving the original face detector's behavior as much as possible. This could be a good design for a personalized face detector where we can train a specialized confidence refinement module for each scenario, but it is out of scope for this paper.

To simplify the design, we take the predicted box and confidence from a face detector as an Nx5 matrix and feed them together with an image into our network. It is possible to further simplify by only refining the confidence without the image input, but this is overfitting to a specific detector. We do not take the feature from the detector's backbone due to the design complication. Choosing the feature is not trivial because of the diversity of network architectures, for example, there are multiple feature scales to be chosen from HRNet, or a large expand layer in MobileNet V3 \cite{mobilenetv3} or EfficientDet \cite{efficientdet} will subsequently increase the refine network size due to the large number of channels in the input. We would like our design to be easily plugged into as many detectors as possible.

Our design avoids the insufficient GPU memory issue in order to train on 8K resolution images. We attempt to resize WiderFace training set to 8K resolution, and train RetinaFace \cite{Deng_2020_CVPR} on Nvidia Quadro 8000 (48GB memory), but this is still not enough for 8K images. Forwarding an 8K image, on the other hand, does not require as much memory, and it is much faster. We forward all the 8K images to obtain 8K boxes and their confidences, then use them to train our confidence ranking network with the original WiderFace resolution. This allows our network to indirectly learn from the 8K images through the 8K prediction of the face detector, and we do not back-propagate to the face detector, so we do not require the large GPU memory.

\subsection{Confidence Ranking Network}
Confidence refinement problem is a sub-task of the detection problem where the region proposal is fixed, and the network learns to predict new confidences. We follow a popular detector design, feature pyramid network (FPN), to inherit the scale variation feature. We then add a box processing network (BPN) by expanding the channel size to 64 (expand Nx5 to 64xNx5) and pass the box feature through multiple convolutions and leaky ReLU layers. We use the convolution with 5x3 kernel to guarantee enough field of view to cover the face detector output (5 numbers: x, y, w, h, confidence). We add paddings to keep the 64xNx5 dimension throughout BPN and we add skip connections between them to preserve the original face detector information. We only interpolate these features with the nearest neighbor before concatenating with the image feature from FPN. The concatenated features are then passed into the remaining FPN layers, and we modify SSH \cite{najibi2017ssh} from a detection module to a confidence module. We first use global average pooling to force the feature dimension to 256 then apply fully connected layers to change the output dimension to the number of confidence from the face detector. We then average all the features across all the scales from FPN to obtain the confidence residual and we sum it back with the predicted confidence from the face detector. We found that learning the new confidence directly (without using the residual) is possible, but the training is less stable, and it takes longer to train. The newly refined confidences are then passed into the ranking loss in Eq.\ref{eq2} during the training. During the testing, we pass the refined confidences and the boxes from the face detector into NMS to obtain the result. Fig. \ref{fig:architecture} demonstrates our pipeline.


\begin{figure}[t]
\centering
\includegraphics[width=0.47\textwidth]{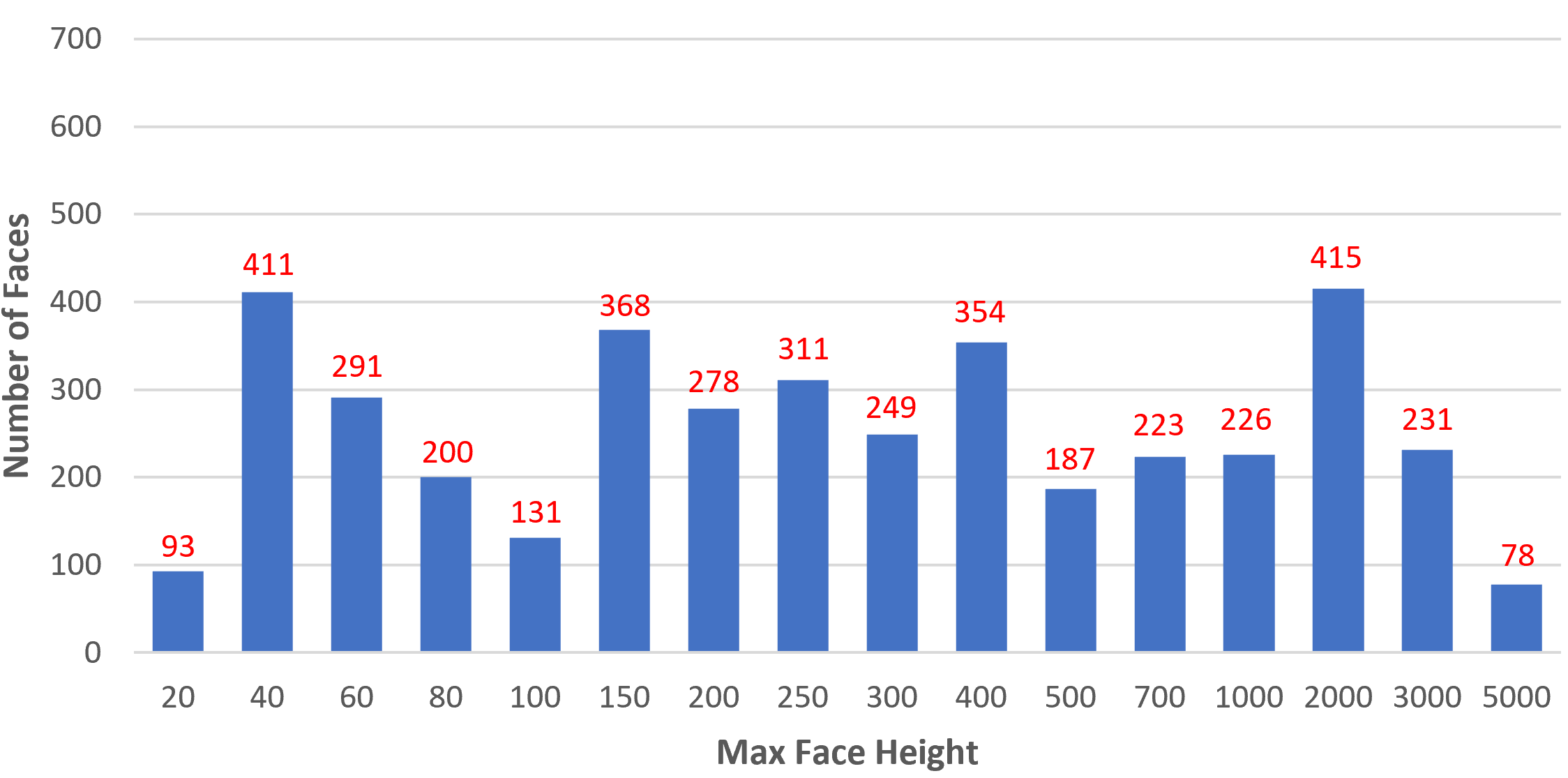}
\caption{Face height distribution of our 8K resolution test set. The X-axis denotes the size range, for example, 40 is the face with $20 <$ height $<= 40$. We cover a wide range of face resolution and has a good balance of the number of faces in each resolution bin.}
\label{fig:8k_dist}
\end{figure}

\section{Experimental Setup}
\subsection{Datasets}
We train our network on WiderFace \cite{yang2016wider} training set and test on WiderFace validation set. WiderFace has 32,203 images with an image width of 1024. This dataset focuses on small faces, where half of all faces are $<=$50 pixels, and there are 393,703 labeled faces (on average: $>$10 faces/image). We collect our 8K resolution test set from Div8K \cite{9021973} and YouTube. We remove the duplicate frames by computing image similarity with the threshold greater 50\%, which ends up with 1,428 images and they cover the full spectrum of face heights from 9 pixels to 4,130 pixels. Fig. \ref{fig:8k_example} shows example images, and Fig. \ref{fig:8k_dist} shows the face size distribution of our 8K resolution test set.

\subsection{Training Configurations}
Our network is implemented in PyTorch \cite{NEURIPS2019_9015} and trained using Adam optimizer with batch size 32 for 100K iterations. The initial learning rate is $10^{-3}$ and it decays exponentially until $10^{-6}$. Training takes about a day on an Nvidia RTX 2080. For data augmentation, we randomly crop, scale, mirror, and change brightness, contrast, hue, and saturation. When we transform the image, we use the same transformation on the predicted box from the face detector directly, without rerunning the augmented image. This is not perfectly accurate, but it is very close and much faster. We normalize the predicted box by the image width and height to keep the input range between 0 to 1. We also clip the refined confidence ($c'$) to be within 0 to 1 to keep the same output range as the standard confidence probability. We use 5K face detection outputs (N=5000). We sort the boxes descendingly by face detector's confidences and remove the boxes if the number exceeds 5K, or we fill the matrix with 0 if there are less than 5K boxes. We combine face detector outputs from RetinaFace, HAMBox, and HRNet, then randomly pick the output from a single face detector on each iteration. The same confidence ranker with NMS (IOU 0.4) is used across all detectors during testing.

\begin{table}
\centering
\begin{tabular}{lccc} 
\hline
\multirow{2}{*}{ \textbf{Loss}} & \multicolumn{3}{c}{\textbf{AP (\%)}}         \\ 
\cline{2-4}
                                & \textbf{Easy} & \textbf{Med} & \textbf{Hard}  \\ 
\hline
Baseline                              & 95.8              & 95.2              & 91.5               \\ 
\hline
Cross Entropy                   & 95.0              & 94.0             & 89.7               \\ 
\hline
L1                              & 95.1              & 94.4             & 90.0               \\ 
\hline
L2                              & 95.1              & 94.3             & 90.4               \\ 
\hline
Smooth L1                       & 95.2              & 94.5             & 90.7               \\ 
\hline
\textbf{Rank}                            & \textbf{96.9}              & \textbf{96.2}             & \textbf{92.7}               \\
\hline
\end{tabular}
\caption{Ablation study on each loss function. The baseline is HRNet on a single-scale. Regression losses are worse than the baseline, but our ranking loss can improve AP Hard by +1.2\%.}
\label{tab:loss}
\end{table}

\begin{table}
\centering
\begin{tabular}{lccc} 
\hline
\multirow{2}{*}{ \textbf{Pair}} & \multicolumn{3}{c}{\textbf{AP (\%)}}         \\ 
\cline{2-4}
                                & \textbf{Easy} & \textbf{Med} & \textbf{Hard}  \\ 
\hline
Baseline                        & 95.8              & 95.2             & 91.5               \\ 
\hline
1                               & 96.7              & 95.8             & 91.9                \\ 
\hline
2                               & 96.8              & 95.9             & 92.2                \\ 
\hline
3                               & 96.7              & 95.9             & 92.1                \\ 
\hline
5                               & 96.8              & 96.0             & 92.3                \\
\hline
\textbf{10}                     & \textbf{96.9}     & \textbf{96.2}    & \textbf{92.7}       \\
\hline
20                              & 96.8              & 96.0             & 92.4                \\
\hline
50                              & 96.8              & 95.9             & 92.1                \\
\hline
100                             & 96.7              & 95.8             & 91.9                \\
\hline
\end{tabular}
\caption{Ablation study on the number of pairs for pairwise ranking loss. A single pair only includes hard pairs, and 100-pairs focuses on easy pairs. Ten pairs yield the best trade-off.}
\label{tab:pair}
\end{table}

\subsection{Network Architecture}
We follow FPN design to extract the image feature with ResNet-50 backbone. BPN consists of 9 convolutions with 64 channels and kernel size 3x5 with padding 1x2 to preserve the same Nx5 dimensions throughout the entire network. We add skip connections on every layer except the first layer. Every 3 layers, we interpolate BPN's features into the image feature dimension and concatenate with the intermediate features from the backbone. The concatenated features are then passed into FPN layers which is a 3x3 convolution with 256 channels and perform the merge on each scale. Each of the FPN features is then passed into a 3x3 convolution with 256 channels, global average pooling, and fully-connected layers (FC) to change into 5000 numbers. FC outputs are then summed up with the face detector's confidence (c) to output the refined confidence ($c'$).

\begin{table}
\centering
\begin{tabular}{lccccc} 
\hline
\multirow{2}{*}{ \textbf{Model}} & \multirow{2}{*}{ \textbf{Scale}} & \multicolumn{3}{c}{\textbf{AP (\%)}}        & \textbf{Speed}  \\ 
\cline{3-5}
                                 &                                  & \textbf{Easy} & \textbf{Med} & \textbf{Hard} & \textbf{(ms)}   \\ 
\hline
\multirow{3}{*}{RetinaFace}      & single                           & 95.2              & 94.5             & 84.3              & 52                 \\ 
\cline{2-6}
                                 & multi                            & 96.4              & 95.5             & 90.3              & 1048            \\ 
\cline{2-6}
                                 & rank                             & 96.1              & 95.3             & 88.7              & 107                 \\ 
\hline
\multirow{3}{*}{HAMBox}          & single                           & 95.1              & 94.2             & 89.1              & 70                \\ 
\cline{2-6}
                                 & multi                            & 96.2              & 95.3             & 90.9              & 1383               \\ 
\cline{2-6}
                                 & rank                             & 96.1              & 95.4             & 91.7              &  126               \\ 
\hline
\multirow{3}{*}{HRNet (ours)}    & single                           & 95.8              & 95.2             & 91.5              & 91               \\ 
\cline{2-6}
                                 & multi                            & 96.9              & 96.1             & 92.1              & 1886                 \\ 
\cline{2-6}
                                 & \textbf{rank}                    & \textbf{96.9}     & \textbf{96.2}    & \textbf{92.7}     & \textbf{149}                 \\ 
\hline
\end{tabular}
\caption{Results on WiderFace validation set. Our confidence ranker (rank) improves AP on top of the single-scale (single) across all backbones. On HRNet, confidence ranker's AP Hard outperforms multi-scale (multi) by +0.6\%, and runs 12.7x faster.}
\label{tab:wider}
\end{table}

\begin{table}
\centering
\begin{tabular}{lccc} 
\hline
\multirow{2}{*}{ \textbf{Model}} & \multirow{2}{*}{ \textbf{Scale}} & \multirow{2}{*}{ \textbf{AP (\%)}} & \textbf{Speed}  \\
                                 &                                  &                                    & \textbf{(ms)}   \\ 
\hline
\multirow{3}{*}{RetinaFace}      & single                           & 24.1                               & 428               \\ 
\cline{2-4}
                                 & multi                            & 38.8                               & 1579                \\ 
\cline{2-4}
                                 & rank                             & 40.7                               & 988                \\ 
\hline
\multirow{3}{*}{HAMBox}          & single                           & 21.8                               & 495                \\
\cline{2-4}
                                 & multi                            & 34.9                               & 1875                \\ 
\cline{2-4}
                                 & rank                             & 37.3                               & 1060                \\ 
\hline
\multirow{4}{*}{HRNet (ours)}    & single                           & 22.5                               & 705                 \\ 
\cline{2-4}
                                 & multi                            & 34.8                               & 2690              \\ 
\cline{2-4}
                                 & rank                             & 37.0                               & 1261                \\
\cline{2-4}
                                 & \textbf{rank 8k}                 & \textbf{42.5}                      & \textbf{1263}                \\
\hline
\end{tabular}
\caption{Results on 8K test set. Our confidence ranker (rank) on the single-scale (scale) outperforms multi-scale (multi) on all backbones, and runs 1.7-2.1x faster. Training confidence ranker on 8K face detection outputs further increases AP by +5.5\%.}
\label{tab:8k}
\end{table}

\section{Results}
\subsection{Ablation Studies}
\noindent\textbf{Importance of Ranking Loss.} We attempt to compare our method with existing detection refinement works, but it is unclear how to change their designs to be model-agnostic, for example, MS R-CNN \cite{huang2019msrcnn} relies on mask head output and RoiAlign features, IoU-Net \cite{Jiang_2018_ECCV} relies on PrRoiPooling features. All these layers do not exist in most face detectors and ignoring these layers will break the refinement because the refiner was trained for these specific features. Loss functions, on the other hand, can be swapped to compare fairly. We use HRNet as the baseline to study the loss functions, and test the AP on WiderFace validation set. Tab. \ref{tab:loss} shows a significant improvement between the ranking loss and all of the other commonly used losses (Cross Entropy is used in \cite{Luo2018,inbook,rongqi,7780745,8794507,7553523}, L2 \cite{huang2019msrcnn}, Smooth L1 \cite{Jiang_2018_ECCV}). The improvement is because the regression losses focus on bringing the confidence magnitude towards the oracle confidence, but the small regression error can easily change the confidence order (as observed in Fig. \ref{fig:pred_dist}). \\


\noindent\textbf{Importance of Pair Selection.} We then tested the number of pairs for our pairwise ranking loss. Since it is too computationally intensive to use all pairs (5K boxes = 12M pairs), we select the pair by sorting the prediction confidence from the face detector and use the neighbor pair. Tab. \ref{tab:pair} shows that when the number of pairs is too small, our network will only focus on the hard pairs. On the other hand, when the number of pairs is too large, the network will only focus on the easy pairs. Empirically, using 10 pairs yields a good trade-off between the hard and easy pairs, and thus achieving the highest AP. We use 10 pairs throughout the paper. We show performance curves in the supplementary.

\subsection{Internal Comparison} \label{sec:internal}
\begin{figure*}[h!]
\centering
\includegraphics[width=1.0\textwidth]{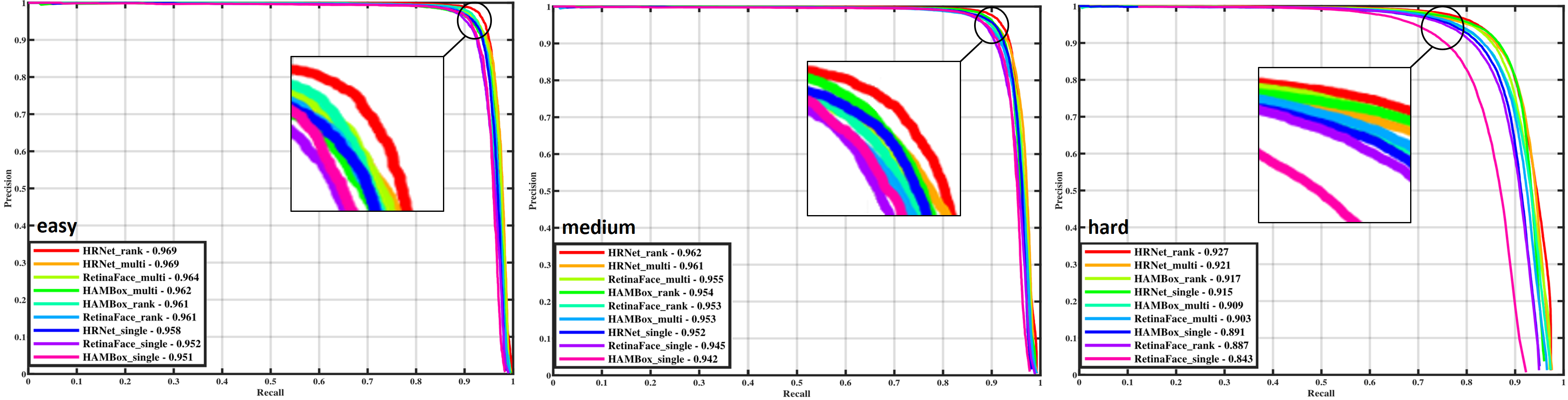}
\caption{Performance curves for internal comparison. We plot the single-scale, multi-scale (multi), and our confidence ranker (rank) across multiple face detectors on WiderFace validation set. HRNet with confidence ranker achieves the highest AP across all settings.}
\label{fig:internal_curve}
\end{figure*}

We show the comparisons of a single-scale, multi-scale, and our confidence ranker on the single-scale on WiderFace validation set. For single-scale testing, we use the image as-is without any resizing (image width is 1024). For multi-scale testing, we follow RetinaFace \cite{Deng_2020_CVPR} by resizing the short edge of the image to [500, 800, 1100, 1400, 1700], as well as flipping the image on each scale, then fuse them together with box voting \cite{Gidaris_2015_ICCV} with an IoU threshold of 0.4. Tab. \ref{tab:wider} shows that on HRNet, our ranking network, despite running on a single-scale, outperforms the AP hard of the multi-scale prediction, and achieves 12.7x speed up. We show the generalization on the open-source RetinaFace \cite{RetinaFaceGithub} and our re-implement HAMBox \cite{HAMBox} in Tab. \ref{tab:wider}. On RetinaFace, the confidence ranker increases AP Hard on the single-scale by 4.4\%, and it is 9.8x faster than multi-scale. On HAMBox, the confidence ranker allows the single-scale to outperform multi-scale on AP hard by 0.8\%, along with 11x speed up. Performance curves are in Fig. \ref{fig:internal_curve}.

\subsection{External Comparison}
\begin{figure*}[h!]
\centering
\includegraphics[width=1.0\textwidth]{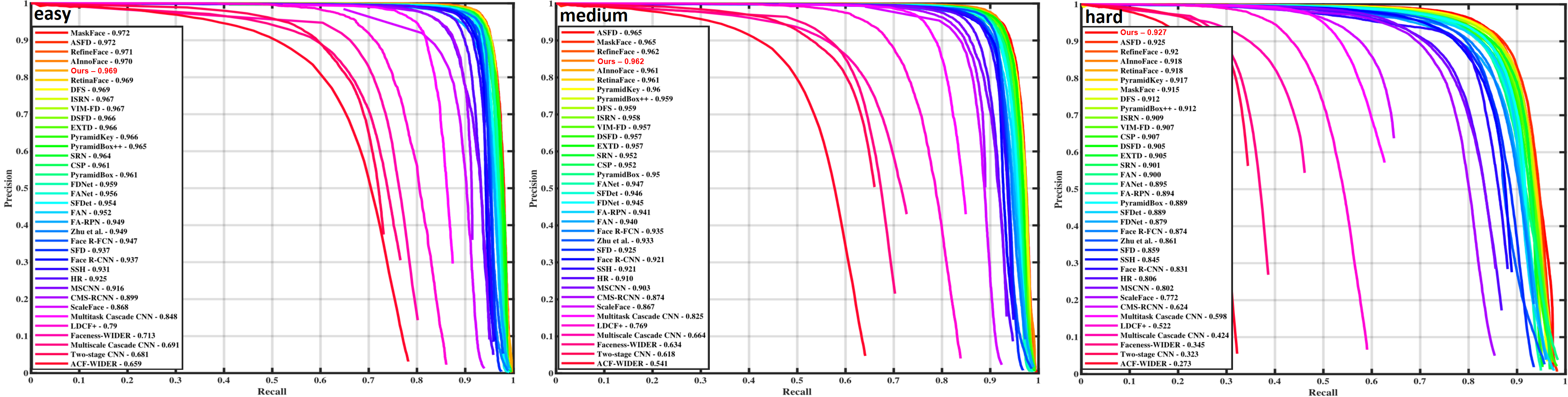}
\caption{Performance curves for external comparison on WiderFace validation set. All of the top 5 detectors (except ours) across all settings use multi-scale. We can approximate multi-scale results with our confidence ranker on a single-scale.}
\label{fig:sot_curve}
\end{figure*}

We externally compare our method on WiderFace validation set. Tab. \ref{tab:sot} shows that we outperform RetinaFace's AP Hard by 0.5\%, and we can estimate HAMBox's AP. We demonstrate in sec.\ref{sec:internal} that our method is compatible with HAMBox, but the AP Hard of our re-implementation is 2.4\% behind HAMBox, so our final model is 0.6\% lower. We show the performance curves for external comparisons in Fig. \ref{fig:sot_curve}. Most of the previous works adopt multi-scale which is significantly slower than the single-scale. RetinaFace reported the unofficial single-scale AP \cite{RetinaFaceGithub} as Easy: 96.5\%, Medium: 95.6\%, Hard: 90.4\%. We outperform RetinaFace by Easy: +0.4\%, Medium: +0.6\%, Hard:+2.3\%, and achieve the highest AP on the single-scale.

\begin{table}
\centering
\begin{tabular}{lcccc} 
\hline
\multirow{2}{*}{\textbf{Model}} & \multicolumn{3}{c}{\textbf{AP (\%) }}    & \textbf{Multi-}           \\ 
\cline{2-4}
                                & \textbf{Easy} & \textbf{Med} & \textbf{Hard}  & \textbf{scale}        \\ 
\hline
S$^3$FD \cite{zhang2017s3fd}                            & 92.8          & 91.3         & 84.0           & Yes  \\ 
\hline
SSH \cite{najibi2017ssh}                            & 92.7          & 91.5         & 84.4     & No       \\ 
\hline
PyramidBox \cite{tang2018pyramidbox}                      & 95.6          & 94.6         & 88.7     & Yes      \\ 
\hline
FA-RPN \cite{Najibi_2019_CVPR}                          & 95.0            & 94.2         & 88.9   & Yes        \\ 
\hline
DSFD \cite{li2019dsfd}                           & 96.0            & 95.3         & 90.0     & Yes        \\ 
\hline
SRN \cite{chi2019selective}                             & 96.4          & 95.3         & 90.2   & Yes         \\ 
\hline
VIM-FD* \cite{vim-fd}                        & 96.7          & 95.7         & 90.7    & -       \\ 
\hline
PyramidBox++* \cite{li2019pyramidbox++}                   & 96.5          & 95.9         & 91.2     & Yes      \\ 
\hline
MaskFace* \cite{yashunin2020maskface}                       & 97.2          & 96.5         & 91.5  & Yes         \\ 
\hline
BFBox \cite{Liu_2020_CVPR}                       & 96.5          & 95.7         & 91.7   & -        \\ 
\hline
AInnoFace* \cite{AInnoFace}                      & 97.0            & 96.1         & 91.8   & Yes        \\ 
\hline
ProgressFace \cite{zhuprogressface}                   & 96.8          & 96.2         & 91.8 & Yes          \\ 
\hline
RefineFace \cite{9099607}                     & 97.1          & 96.2         & 91.8  & Yes          \\ 
\hline
ASFD* \cite{zhang2020asfd}                          & 97.2          & 96.2         & 92.0   & Yes          \\ 
\hline
RetinaFace \cite{Deng_2020_CVPR}                     & 96.9          & 96.3         & 92.2  & Yes         \\ 
\hline
HAMBox \cite{HAMBox}                         & 97.0            & 96.4         & 93.3     & Yes      \\ 
\hline
\textbf{Ours}                  & \textbf{96.9}              & \textbf{96.2}             & \textbf{92.7}          &   \textbf{No}  \\
\hline
\end{tabular}
\caption{External comparison on WiderFace validation set. Works with * are not formally published. Our method is competitive with multi-scale, and achieve the highest AP on single-scale.}
\label{tab:sot}
\end{table}

\begin{figure*}[h!]
\centering
\includegraphics[width=1.0\textwidth]{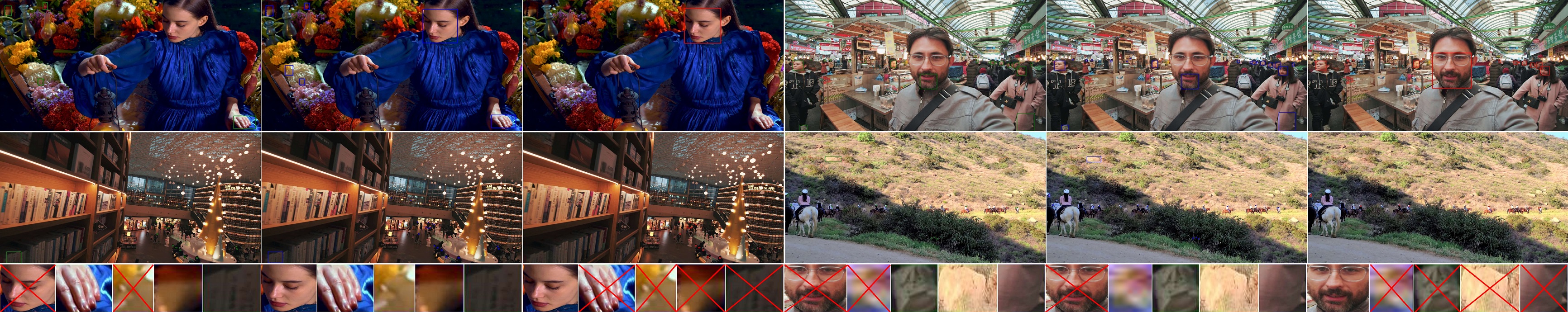}
\caption{(Best viewed electronically) Prediction examples on 8K test set. We show the predictions from HRNet with confidences $>$0.5 on the single-scale (left), multi-scale (mid), and confidence ranker trained on 8K (right). Single-scale misses multiple large faces, and multi-scale is more prone to false positives. Our confidence ranker can mitigate both failure cases.}
\label{fig:8k_predict_example}
\end{figure*}

\subsection{8K Resolution}
We study the impact of single-scale, multi-scale, and our method on 8K resolution test set. For multi-scale, we follow RetinaFace by down-sampling the smallest scale to approximately half of the original image dimension. We omit the up-sampling scales because the 48GB GPU memory is not enough to run them. We resize the long edge to [7680, 5760, 4096] and flip the image on each scale, then fuse them together with box voting. Tab. \ref{tab:8k} shows that our confidence ranker on the single-scale outperforms the multi-scale on all three detectors. The AP improvements are +1.9\%, +2.4\%, +2.2\%, and the speedups are 1.7x, 1.8x, 2.1x for RetinaFace, HAMBox and HRNet respectively. This shows the generalization of our method. By resizing the WiderFace training set to 8K resolution and feed them into all three detectors, we can train our confidence ranker on 8K face detector outputs (which is not possible to train on the face detector directly due to GPU memory limitation). Our confidence ranker, trained on WiderFace in 8K resolution, can further increase AP on HRNet by 5.5\% and the speed remains unchanged. We show the visual comparison in Fig. \ref{fig:8k_predict_example} and the performance curve in the supplementary.


\subsection{Computational Complexity}
Our confidence ranker is 128MB and the speed on an Nvidia RTX 2080, is 45ms/frame and 550ms/frame for 1024 and 8K respectively. Our confidence ranker is in this size because face detectors are the bottleneck where HRNet and HAMBox run at 81ms/frame and 65ms/frame on 1024. To achieve a smaller footprint, we change the backbone from ResNet-50 to MobileNet v1 \cite{howard2017mobilenets} and reduce the channel size of the confidence module from 256 to 64 (this will reduce the FC layer from 256$=>$5000 to 64$=>$5000). Our small model is 8MB and it achieves AP Easy: 96.6\%, Medium: 95.7\%, Hard: 91.7\% on WiderFace validation set with 6ms/frame on a 1024 and 59ms/frame on 8K.
\section{Conclusion and Future Works}
We propose a confidence ranking network with a pairwise ranking loss to re-rank the incorrect confidences. On WiderFace, we achieve the highest AP on the single-scale, and we can approximate the AP of the previous multi-scale methods with a few times speed up. Our confidence ranker is model-agnostic and supports training indirectly on 8K resolution. For future work, it is possible to combine ranking loss with other techniques, such as adversarial loss, for improving challenging cases such as blurry, underexposed, and heavily occluded faces. In theory, our approach can also be applied for generic object detection and instance segmentation. We leave them as future explorations.



\clearpage
{\small
\bibliographystyle{splncs04}

}
\clearpage








\title{\elvbfx{CRFace: Confidence Ranker for Model-Agnostic Face Detection Refinement} \\ \Large{(Supplementary)}}

\author{Noranart Vesdapunt\\
Microsoft Cloud\&AI	\\
{\tt\small noves@microsoft.com}
\and
Baoyuan Wang\\
Xiaobing.AI\\
{\tt\small zjuwby@gmail.com	}
}
\date{\vspace{-3ex}}

\maketitle

We propose a confidence ranking network with a pairwise ranking loss to re-rank the predicted confidences locally within the same image to improve average precision (AP) across multiple face detectors on WiderFace \cite{yang2016wider} and our 8K resolution test set. In this supplementary, we provide performance curves for section 5.1 and 5.4, and additional visualization results from our 8K resolution test set.

\section{Performance Curves}
\begin{figure*}[h!]
\centering
\includegraphics[width=0.95\textwidth]{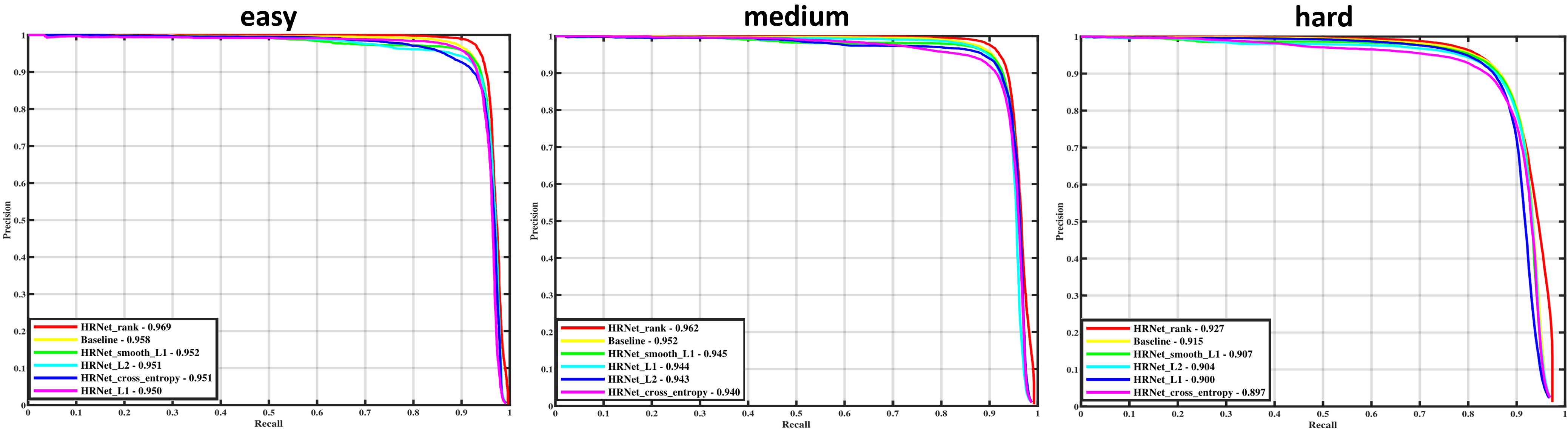}
\caption{Performance curves for each loss function. Regression losses decrease AP due to the focus on confidence magnitude rather than confidence order. Our ranking loss improves AP Hard by +1.2\%.}
\label{fig:loss}
\end{figure*}

\begin{figure*}[h!]
\centering
\includegraphics[width=0.95\textwidth]{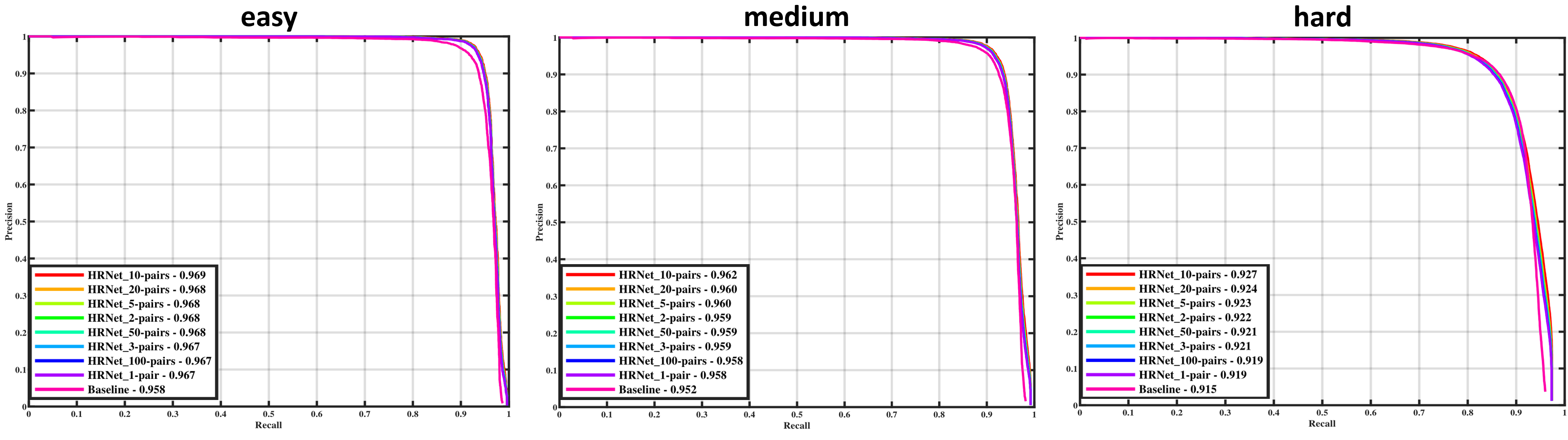}
\caption{Performance curves for each number of pairs for pairwise ranking loss. Choosing a single pair will omit easy pairs, and choosing 100-pairs will disregard hard pairs. 10-pairs provides the best trade-off between easy and hard pairs.}
\label{fig:pair}
\end{figure*}

\noindent\textbf{Ablation Study.} We show performance curves for both of our ablation studies in section 5.1 on WiderFace validation set \cite{yang2016wider}. Our baseline is a re-implementation of HAMBox \cite{HAMBox} and we replace ResNet-50 backbone with HRNet \cite{WangSCJDZLMTWLX19,SunXLW19}. We show the importance of ranking loss in Fig. \ref{fig:loss} where the regression losses are struggling to preserve the confidence order and result in an even worse AP than the baseline, while our ranking loss improves AP by 1-1.2\%. The importance of pair selection is shown in Fig. \ref{fig:pair} where 10-pairs provides the best trade-off between the easy and hard pairs, thus achieving the highest AP. \\

\begin{figure}[h]
\centering
\includegraphics[width=0.35\textwidth]{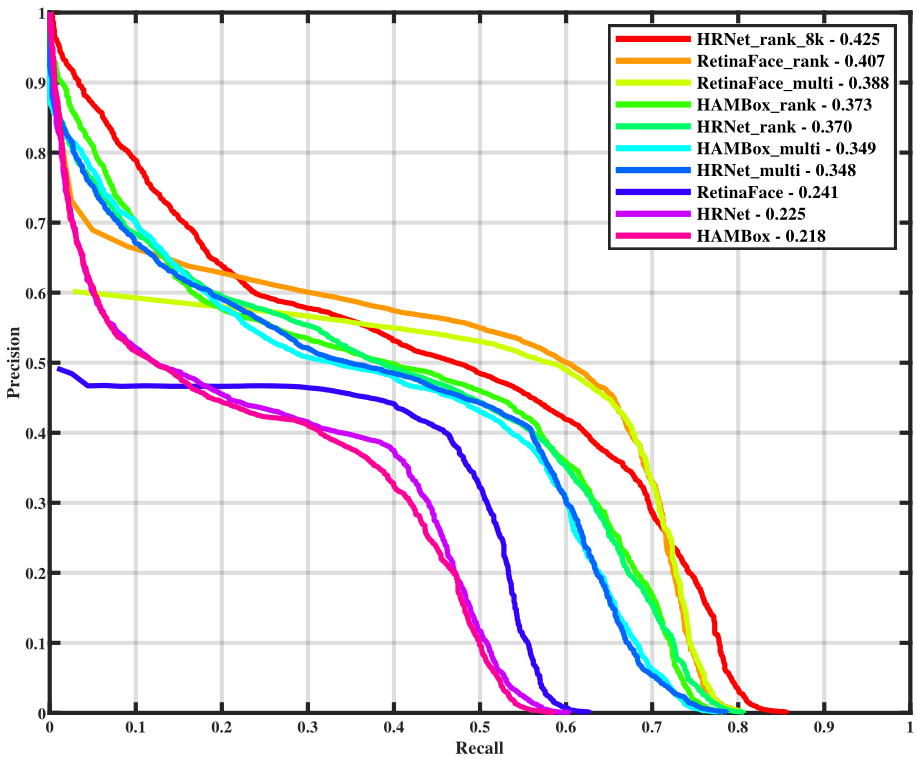}
\caption{Performance curves on 8K test set. Our confidence ranker (rank) on a single-scale (scale) has higher AP than multi-scale (multi) on all backbones and runs 1.7-2.1x faster.}
\label{fig:8k}
\end{figure}

\noindent\textbf{8K Resolution.} We show the performance curves of the single-scale, multi-scale, and our method on 8K resolution test set in Fig. \ref{fig:8k}. Our confidence ranker is model-agnostic, and it can improve AP across all three detectors (RetinaFace \cite{Deng_2020_CVPR}, HAMBox, HRNet), and allows us to train indirectly on 8K resolution to further boost AP without any speed decrease during test time.

\section{Visualization of 8K Test Set}
\begin{figure*}[h!]
\centering
\includegraphics[width=1.0\textwidth]{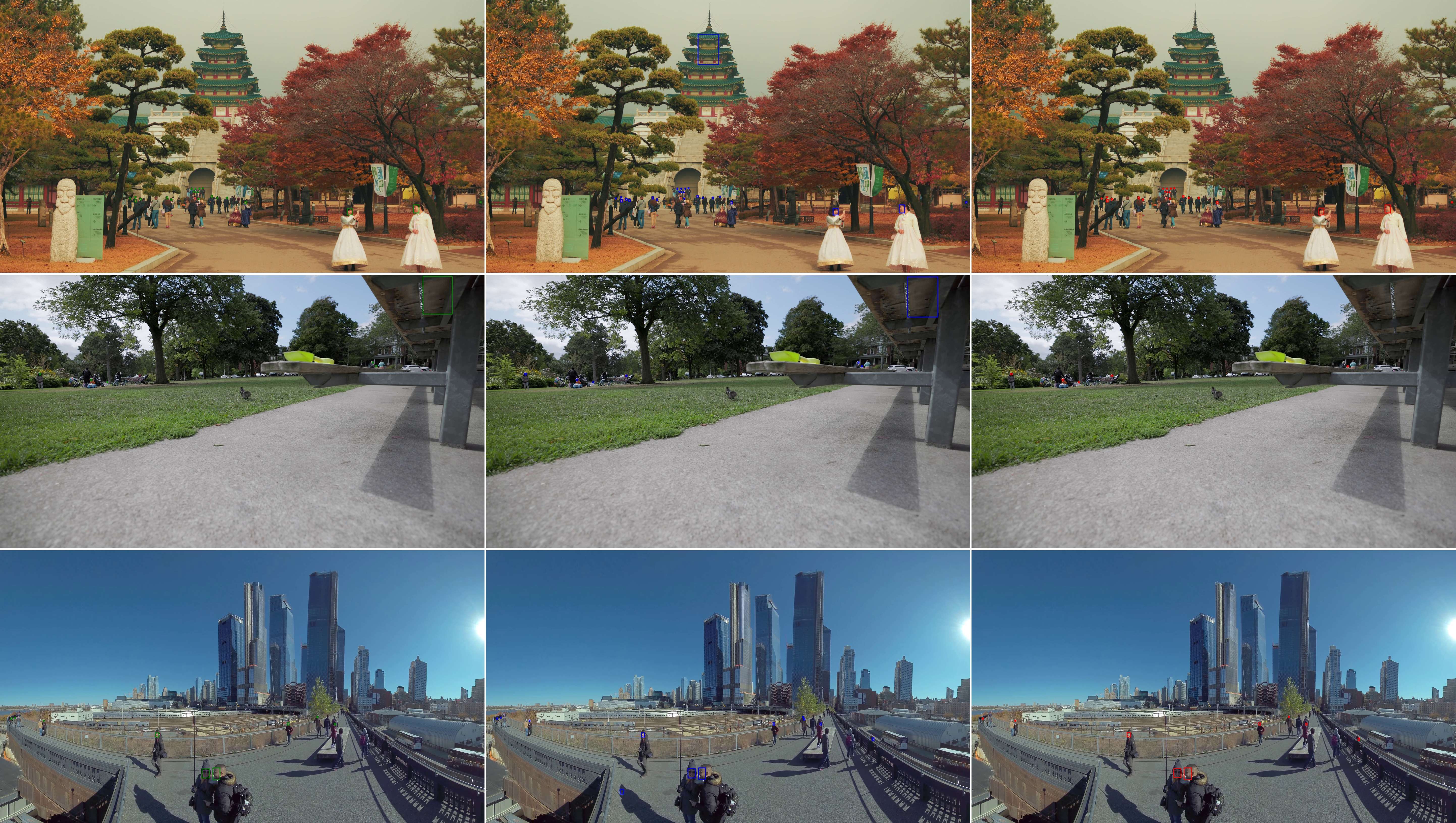}
\caption{(Best viewed electronically) Prediction examples on 8K test set. We show the predictions from HRNet with confidences $>$0.5 on the single-scale (left), multi-scale (mid), and confidence ranker trained on 8K (right).}
\label{fig:sample1}
\end{figure*}

\begin{figure*}[h!]
\centering
\includegraphics[width=1.0\textwidth]{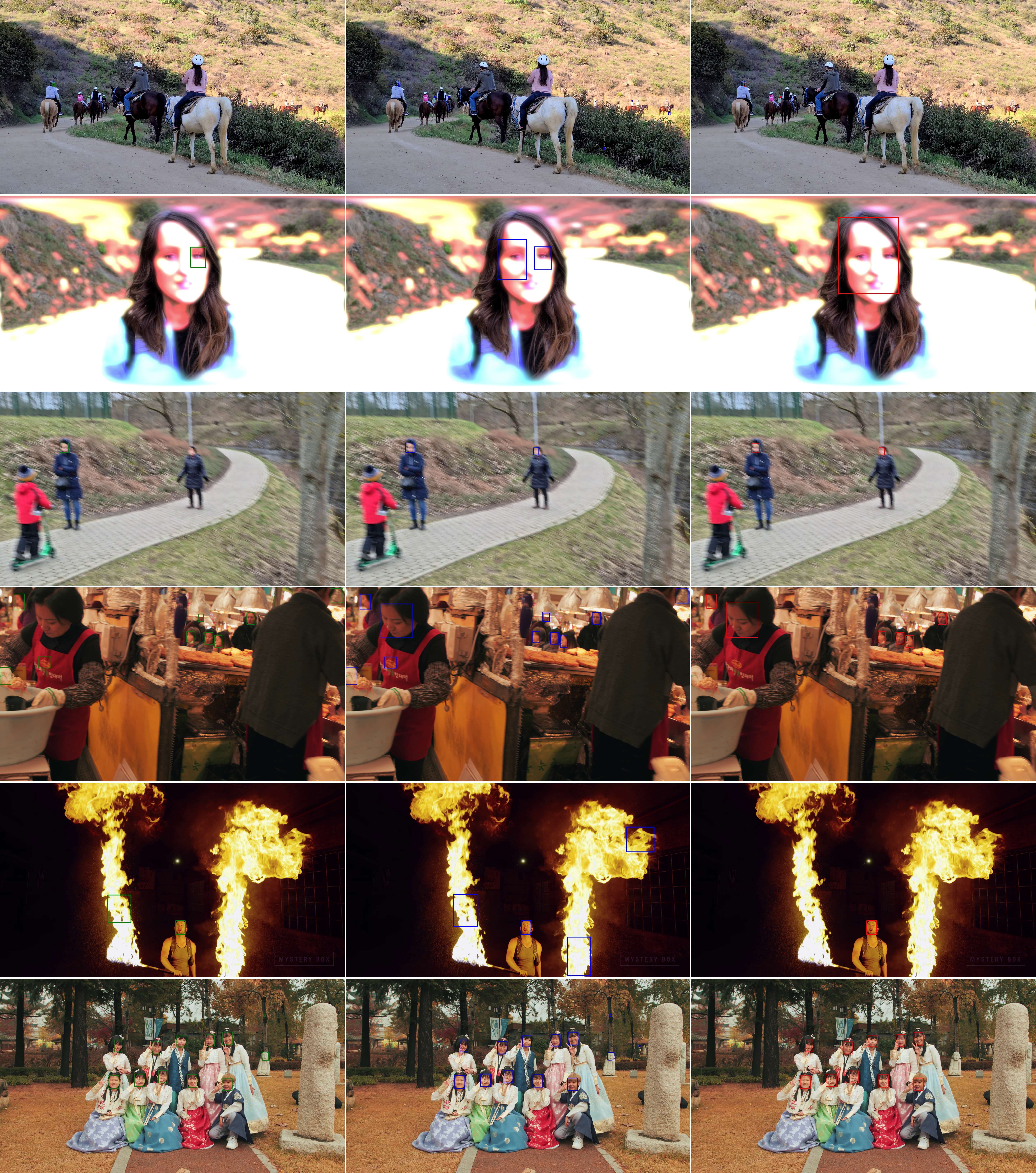}
\caption{(Best viewed electronically) Prediction examples on 8K test set. We show the predictions from HRNet with confidences $>$0.5 on the single-scale (left), multi-scale (mid), and confidence ranker trained on 8K (right).}
\label{fig:sample2}
\end{figure*}

\begin{figure*}[h!]
\centering
\includegraphics[width=1.0\textwidth]{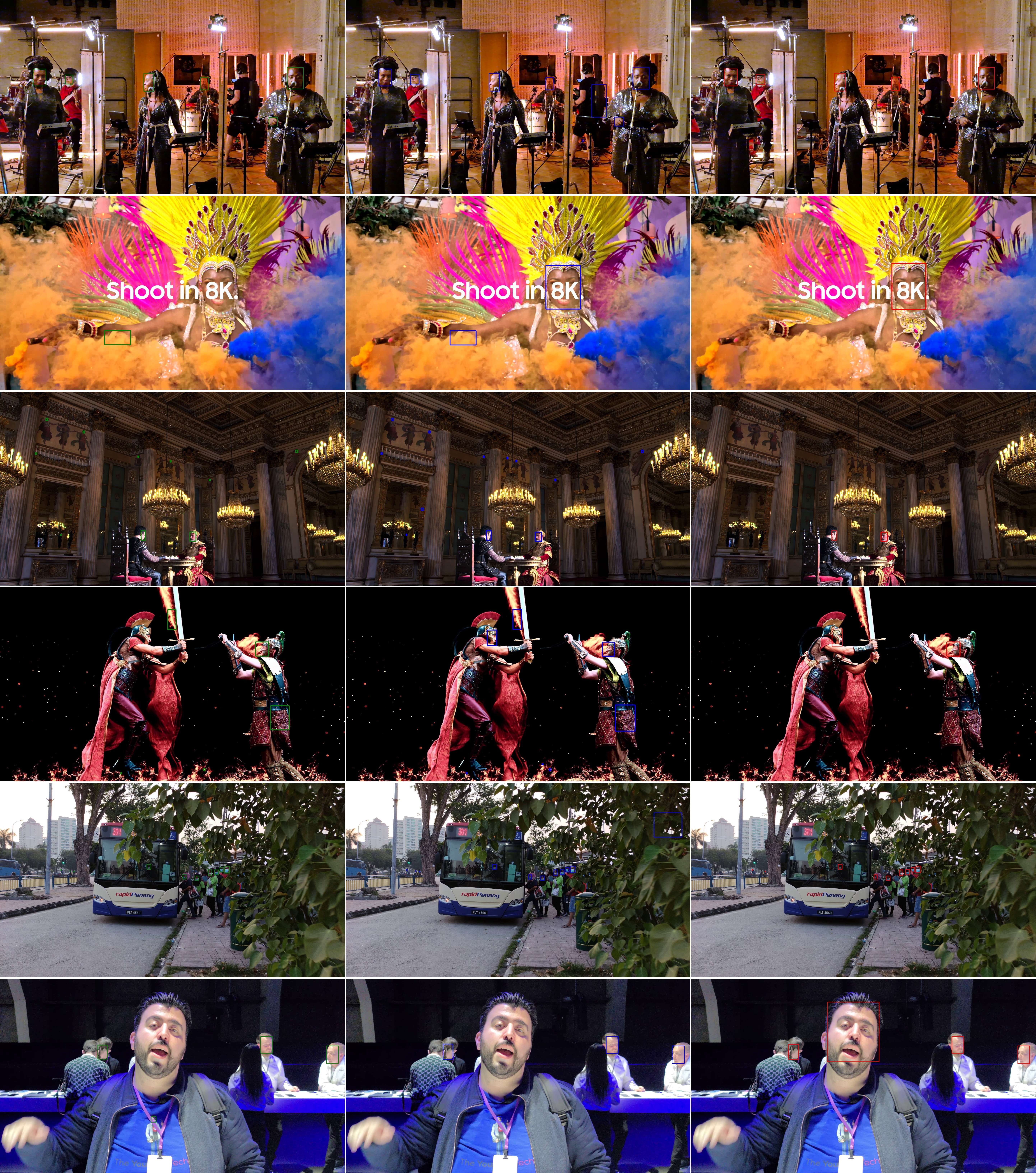}
\caption{(Best viewed electronically) Prediction examples on 8K test set. We show the predictions from HRNet with confidences $>$0.5 on the single-scale (left), multi-scale (mid), and confidence ranker trained on 8K (right).}
\label{fig:sample3}
\end{figure*}

We show more visual comparisons between the single-scale, multi-scale, and our method on 8K resolution test set in Fig.\ref{fig:sample1},\ref{fig:sample2},\ref{fig:sample3}. In general, single-scale usually fails to predict large face due to the maximum anchor size is only 512 and the training data only has an image width of 1024. Multi-scale forwards the images 6 times on 3 resolutions and flip on each scale, so the smaller resolution performs a better job at predicting large face, but it also generates more false positive. Our method improves on top of single-scale, so we predict less false positive than multi-scale, and achieves the best result while running 1.7-2.1x faster.

{\small
\bibliographystyle{splncs04}

}


\end{document}